\documentclass{article}
\usepackage[dblblindworkshop,final]{neurips_2026}
\workshoptitle{Linguistic Principles for Foundation Models}
\usepackage[utf8]{inputenc}
\usepackage[T1]{fontenc}
\usepackage{float}
\usepackage{capt-of}
\usepackage{hyperref}
\usepackage{url}
\usepackage{booktabs}
\usepackage{amsmath,amsfonts}
\usepackage{microtype}
\usepackage{xcolor}
\usepackage{graphicx}
\usepackage{multirow}
\usepackage{array}
\usepackage{placeins}
\hypersetup{colorlinks=true,allcolors=blue!45!black}
\graphicspath{{figures/}}
\title{Table Serialization Determines\\
Which Relations Language Models Preserve}

\author{
  {\small
  Junwei Ma$^{1}$\thanks{Corresponding author: \texttt{junweima2@gmail.com}} \quad
  Arun Kumar Chithanar \quad
  Palash Goyal$^{2}$ \quad
  Chul Lee
  }\\[4pt]
  {\footnotesize
  $^{1}$University of Toronto \quad $^{2}$Google
  }
}
\begin{document}
\maketitle

\begin{abstract}
We study how text serialization affects table lookup in instruction-tuned
language models. Our diagnostic asks the model to find the row containing a
specified value under one named column, then return the value under another
named column in that same row. We render the same synthetic 256-cell tables in row-major form
and in column-major form with each header placed beside its values, while
varying the table aspect ratio across five shapes. Across three models, the
better format changes with table shape: row-major performs better on tall
tables, whereas column-major performs better on wide tables. At $64\times4$,
row-major leads by 42.3--54.0 percentage points; at $4\times64$, column-major
leads by 35.0--59.0 points. Mapping wrong answers back to their source cells
reveals a complementary error pattern: row-major tends to preserve the correct
row but lose the target column, while column-major tends to preserve the
correct column but lose the row. We use the term \emph{relational locality} to
describe this pattern: serialization keeps one relation locally available
while making the other depend on positional alignment. A separate non-tabular
ordering experiment supports this interpretation.
Thus, apparent width-related failures in text-serialized tables depend on
linearization, not table shape alone.
\end{abstract}

\section{Introduction}
Tables are a common way to organize information in spreadsheets, databases,
reports, and software tools. General-purpose language models are increasingly
asked to answer questions over such data
\citep{sui2023table,liu2024rethinking,jiang2023structgpt}. When a table is
included in a text prompt, however, it must be written as a sequence---usually
row by row or column by column. These renderings can contain exactly the same
cells while placing different labels and values next to one another.

Under the common row-major format, accuracy on some table tasks falls as more
columns are added. This observation alone does not show that wide tables are
inherently difficult: adding columns also lengthens every serialized row and
increases the distance between the header and its values. Table shape and
serialization are therefore confounded. A controlled comparison must hold the
table itself fixed while changing how it is written.

We make this comparison using a simple lookup task. The model is told that a
particular value appears in one named column and must return the value from a
different named column in the same row. We render the same synthetic
256-cell tables in row-major and locally headed column-major form, vary their
aspect ratio across five shapes, and evaluate three instruction-tuned language
models. Because the cell contents, question, and answer are unchanged within
each pair, differences in accuracy can be attributed to the representation.

The better representation changes with table shape. Row-major text performs
better on tall tables, whereas column-major text performs better on wide
tables. The errors also reverse: row-major predictions more often retain the
correct row but miss the requested column, while column-major predictions more
often retain the correct column but miss the row. We call this pattern
\emph{relational locality}: a serialization makes some relations easier to
recover by keeping their relevant values and labels together. This motivates a
simple hypothesis: language models recover a relation more reliably when the
serialization keeps those elements together. We test it through a paired
row-versus-column lookup experiment and a separate non-tabular ordering
experiment. Our main conclusion is that apparent width-related failures in
text-serialized tables depend on linearization as well as table shape.

\section{Experiments}
\subsection{Row-major versus column-major lookup}

\paragraph{Task.}
Each synthetic table has named columns and unique values. The question
identifies a row by giving a known value in one column, then asks for the value
in a different column from that same row.
Let $T\in\mathcal{V}^{R\times C}$ contain unique values. A query supplies a
predicate $(c_p,v_p)$ that determines row $r$ through $T_{r,c_p}=v_p$, and a
different target column $c_t$. The answer is $T_{r,c_t}$. Crossover prompts add
no row labels, although their predicate values use a distinct type.

The lookup is semantically equivalent to the following database query (we do
not execute SQL):
\begin{quote}
\ttfamily\small
SELECT Salary\\
FROM employees\\
WHERE Employee = 'Bob';
\end{quote}
Figure~\ref{fig:prompt-example} shows the two natural text renderings used in
the main comparison. All paired prompts contain exactly the same logical
table, display-column order, question, and answer.

\FloatBarrier
\begin{figure}[H]
\centering
\scriptsize
\setlength{\tabcolsep}{3pt}
\begin{tabular}{>{\raggedright\arraybackslash}p{.47\linewidth}>{\raggedright\arraybackslash}p{.47\linewidth}}
\toprule
Row-major & Column-major with local headers \\
\midrule
\texttt{Column order: Employee | Department | Salary}\newline
\texttt{Grouping: rows}\newline
\texttt{Alice | Sales | 72000}\newline
\texttt{Bob | Engineering | 95000}\newline
\texttt{Carla | Finance | 83000}
&
\texttt{Grouping: columns}\newline
\texttt{Column Employee: Alice | Bob | Carla}\newline
\texttt{Column Department: Sales | Engineering | Finance}\newline
\texttt{Column Salary: 72000 | 95000 | 83000}
\\
\bottomrule
\end{tabular}
\vspace{3pt}
\begin{minipage}{.96\linewidth}
\textbf{Question:} What is the value in column \texttt{Salary} for the row
where column \texttt{Employee} equals \texttt{Bob}? Return only the value. Do
not explain your answer. Your entire response cannot exceed 30 tokens.
\textbf{Answer:} \texttt{95000}.
\end{minipage}
\caption{The same logical table in the two main experimental serializations.
Readable names replace the random strings used in the experiments. Neither
representation adds row labels; column-major labels each column once.}
\label{fig:prompt-example}
\end{figure}

\paragraph{Hypothesis.}
The \emph{relational locality} hypothesis predicts that a format will favor the
relation it keeps together. Row-major keeps rows together but requires a
global header-to-position mapping; locally headed column-major keeps each
header beside its values but requires alignment across columns. We therefore
predict that added columns will hurt row-major, whereas added rows will hurt
column-major.

\begin{figure}[H]
\centering
\includegraphics[width=.88\linewidth]{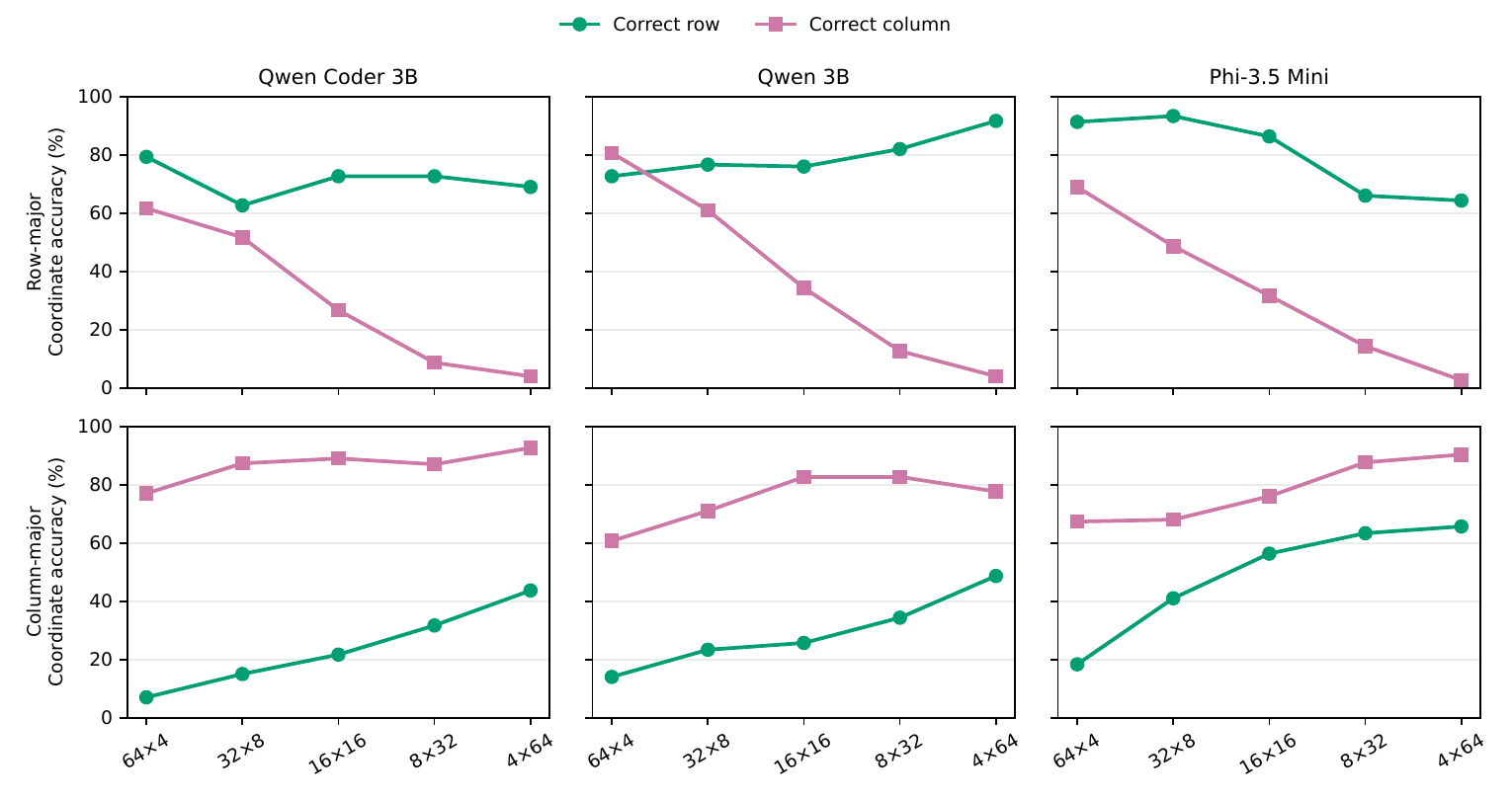}
\caption{Row and column accuracy. Exact answers count toward both lines.
Row-major increasingly loses columns with width; column-major increasingly
loses rows with depth.}
\label{fig:coordinates}
\end{figure}

\paragraph{Protocol.}
We test five 256-cell shapes from $64\!\times\!4$ to $4\!\times\!64$, with 300
paired tables per shape and 3,000 prompts per model. We evaluate two Qwen 3B
models and Phi-3.5 Mini with greedy decoding
\citep{hui2024qwen25coder,yang2024qwen25,abdin2024phi3}. Exact accuracy requires
the target cell; Figure~\ref{fig:coordinates} additionally checks whether a
wrong value comes from the correct row or requested column. Error bars are
Wilson 95\% intervals. Full generation, parsing, and inference details are in
Appendix~\ref{app:protocols}.

\begin{figure}[H]
\centering
\includegraphics[width=.90\linewidth]{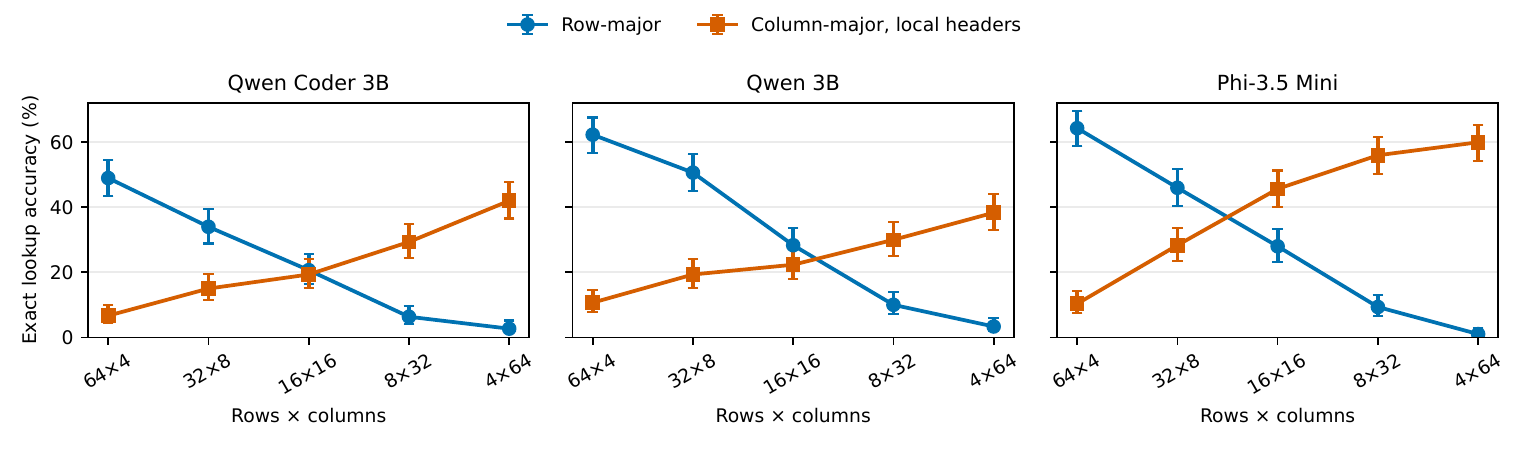}
\caption{Exact lookup accuracy with Wilson 95\% intervals. Each point contains
300 questions paired between row-major and locally headed column-major
representations.}
\label{fig:crossover}
\end{figure}

\paragraph{Accuracy crossover.}
Figure~\ref{fig:crossover} shows the crossover in all three models. Row-major
dominates at the two tall shapes, the curves approach one another near square,
and locally headed column-major dominates at the two wide shapes. At
$64\times4$, column-minus-row differences range from $-42.3$ to $-54.0$
points; at $4\times64$, they range from $+35.0$ to $+59.0$ points
(Appendix Table~\ref{tab:crossover-effects}). The paired differences directly show that
the representation advantage reverses between the tallest and widest shapes,
as relational locality predicts.

Valid-value extraction is 95.3--98.3\% for the Qwen models and 99.0--100\% for
Phi, so output truncation cannot explain the interaction. The predictions are
mostly wrong cells from the displayed table.

\paragraph{Where wrong answers come from.}
Figure~\ref{fig:coordinates} checks whether each valid prediction comes from
the correct row or the requested column. The errors are not undifferentiated
retrieval failures. For Qwen
Coder at $4\times64$, row-major retains the correct row in 69.0\% of prompts
but the correct column in only 4.0\%, producing 2.7\% exact accuracy. At the
transposed aspect ratio, $64\times4$ column-major retains the correct column in
77.0\% but the correct row in only 7.0\%, producing 6.7\% exact accuracy. The
corresponding random-valid-cell baselines are 25\% for the shorter dimension
and 1.6\% for the longer dimension. The same reversal appears in the other
checkpoints. Serialization therefore preserves one dimension while making the
other fragile.
Individual outputs exhibit the same distinction: a wide row-major error can
select another field from the identified record, while a tall column-major
error can select the requested field from a different record.
Together with the crossover, this reversal in where errors come from is the
main evidence for relational locality in table lookup.

Appendix~\ref{app:grid} reports an independent-axis replication and additional
serialization ablations.

\subsection{Non-tabular pairwise intersection: grouped versus interleaved}

\paragraph{Task.}
To test whether relational locality extends beyond table coordinates, we
construct a sequence of labeled \texttt{GROUP=VALUE} records. Every pair of
groups shares exactly one unique value, and the model must identify the value
shared by the two groups named in the question.

\paragraph{Example.}
A simplified three-group instance makes the ordering change concrete:
\begin{quote}
\footnotesize
\textbf{Grouped order:}\quad
\texttt{Red=apple | Red=violin | Red=cloud}\newline
\texttt{Blue=apple | Blue=river | Blue=lamp}\newline
\texttt{Green=violin | Green=river | Green=chair}

\textbf{Two-fragment order:}\quad
\texttt{Red=apple | Red=violin | Blue=apple | Blue=river}\newline
\texttt{Green=violin | Green=river | Red=cloud | Blue=lamp | Green=chair}

\textbf{Interleaved order:}\quad
\texttt{Red=cloud | Blue=apple | Green=violin | Red=violin}\newline
\texttt{Blue=river | Green=river | Red=apple | Blue=lamp | Green=chair}

\textbf{Question:} Which value occurs in both Red and Blue?
\textbf{Answer:} \texttt{apple}.
\end{quote}
All versions contain the same records. A \emph{fragment} is one uninterrupted
group block, so each group has one, two, or three above. In the middle order,
Red's two fragments are \texttt{Red=apple | Red=violin} and
\texttt{Red=cloud}. The experiment uses four groups and longer numeric lists.

\paragraph{Hypothesis.}
The \emph{relational locality} hypothesis predicts that keeping a queried
group together makes its shared value easier to recover, even without table
rows or columns. Accuracy should therefore fall as the same records are split
into more fragments.

\paragraph{Protocol.}
For 100 base instances at each of 8 and 16 values per group, we render 1, 2,
4, 8, or 16 fragments, yielding 900 prompts. Paired conditions keep the
records, tokens, question, and answer fixed. We evaluate
Qwen2.5-Coder-3B-Instruct with greedy decoding and exact-match scoring. Error
bars are Wilson 95\% intervals; contrasts use paired bootstrap intervals. Full
details are in Appendix~\ref{app:protocols}.

\noindent
\begin{minipage}[t]{.54\linewidth}
\vspace{0pt}
\paragraph{Results.}
With 8 values per group, accuracy falls from 36\% with one fragment to 24\%
under full interleaving; with 16, it falls from 33\% to 11\%
(Figure~\ref{fig:locality}). Across lengths, one fragment adds 17.0 points over
full interleaving (95\% paired CI $[9.0,25.0]$), and each fragmentation
doubling costs 5.2 points ($[2.6,7.6]$). Because content is fixed, this supports
relational locality beyond table coordinates.
\end{minipage}\hfill
\begin{minipage}[t]{.41\linewidth}
\vspace{0pt}
\centering
\includegraphics[width=\linewidth]{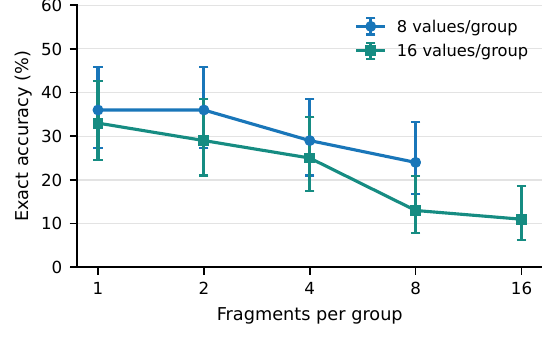}
\captionof{figure}{Accuracy versus fragments per group (Wilson 95\%
intervals).}
\label{fig:locality}
\end{minipage}

\FloatBarrier
\enlargethispage{2\baselineskip}

\vspace{-4pt}
\section{Related work}
Table-aware language models encode structure directly
\citep{herzig2020tapas,yin2020tabert,liu2022tapex}, while tabular foundation
models target supervised numerical prediction
\citep{hollmann2025tabpfn,ma2025tabdpt,qu2025tabicl}. We instead study an
unmodified instruction-tuned model receiving tables as text. Prior work shows
that serialization, structural perturbations, and structured-data interfaces
affect language-model performance
\citep{sui2023table,singha2023tabular,liu2024rethinking,sui2024tap4llm,
jiang2023structgpt,dong2024encoding,li2025longtablebench}, and that retrieval is
position-sensitive \citep{liu2024lost}. Our paired design holds the logical
table and query fixed, traces wrong values to source cells, and tests the same
locality prediction in a non-tabular ordering task.

\vspace{-4pt}
\section{Discussion and limitations}
Together, the experiments support relational locality: models more reliably
recover relations kept close and grouped in the serialized prompt. The
row--column crossover therefore is not evidence that width alone makes a table
difficult: the formats move different relations into local spans, and the
errors move with them. Serialization is part of the model-facing
representation; when text is necessary, the format should keep the
task-relevant relation local. Exact production table retrieval should still
use SQL or dataframes when possible \citep{jiang2023structgpt}. Our evidence uses
synthetic, unique-valued inputs and two narrow diagnostic tasks; only the
crossover spans three models, while the ordering study uses one. We identify
no mechanism, did not preregister, and leave natural tables and other tasks
open.

\clearpage
\bibliographystyle{plainnat}
\bibliography{references}

\appendix
\section{Detailed experimental protocols}
\label{app:protocols}
For lookup, prompts use native chat templates, pinned revisions, bfloat16
weights, greedy decoding, no demonstrations or training, and a 64-token cap.
We extract the first generated six-digit value that occurs in the table. Exact
accuracy requires the target cell; a valid wrong value receives row credit if
it lies in the target row and column credit if it lies in the requested
column. There are no response-based exclusions, and no retained completion
exceeds 10 tokens. Wilson intervals summarize conditions; representation
contrasts use 10,000 paired bootstrap resamples over logical tables and exact
McNemar tests.

For pairwise intersection, each base instance has four groups and one unique
shared value per group pair. The 900 prompts cover 100 instances at each of 8
and 16 values per group; paired versions have identical records, values,
questions, answers, prompt length, and tokenizer multiset.
Qwen2.5-Coder-3B-Instruct uses its native chat template, bfloat16 weights, greedy decoding, no
demonstrations or training, and a 30-token cap. We extract the first generated
six-digit value and require the value shared by the queried groups. There are
no response-based exclusions, no completion exceeds six tokens, and contrasts
use paired bootstrap intervals.

\section{Protocol audit and deviations}
The original confirmatory document prespecified a Mistral-7B checkpoint as the
third model. The executed run instead used Phi-3.5-mini-instruct. We report Phi
as an additional cross-family replication but do not describe the executed
three-model list as preregistered.

The same document prespecified a coordinate-label factorial. A post-run audit
found that it repeated the predicate value beside every cell, effectively
creating a row identifier, and that its column-major form did not preserve the
locally headed grammar. We therefore exclude that factorial from every claim
and figure. Its raw outputs remain in the artifact for provenance. The
replacement independent-axis grid removes the repeated row descriptor but
retains a type-distinguished predicate field, so we report it only as a
replication and ablation rather than as confirmatory evidence.

\section{Additional lookup results}
\label{app:grid}
Table~\ref{tab:crossover-effects} gives the paired numerical contrasts behind
the main crossover figure.

\begin{table}[ht]
\centering
\scriptsize
\caption{Paired column-major-minus-row-major accuracy differences at the tall, square, and wide shapes. Intervals use 10,000 paired bootstrap resamples.}
\label{tab:crossover-effects}
\begin{tabular}{lrrr}
\toprule
Model & $64\times4$ & $16\times16$ & $4\times64$ \\
\midrule
Qwen Coder 3B & -42.3 [-48.7, -36.0] & -1.3 [-7.3, +5.0] & +39.3 [+33.7, +45.0] \\
Qwen 3B & -51.7 [-57.7, -46.0] & -6.0 [-12.7, +0.7] & +35.0 [+29.0, +41.0] \\
Phi-3.5 Mini & -54.0 [-59.7, -48.0] & +17.7 [+10.7, +24.7] & +59.0 [+53.3, +64.7] \\
\bottomrule
\end{tabular}
\end{table}

\FloatBarrier

\begin{figure}[ht]
\centering
\includegraphics[width=.98\linewidth]{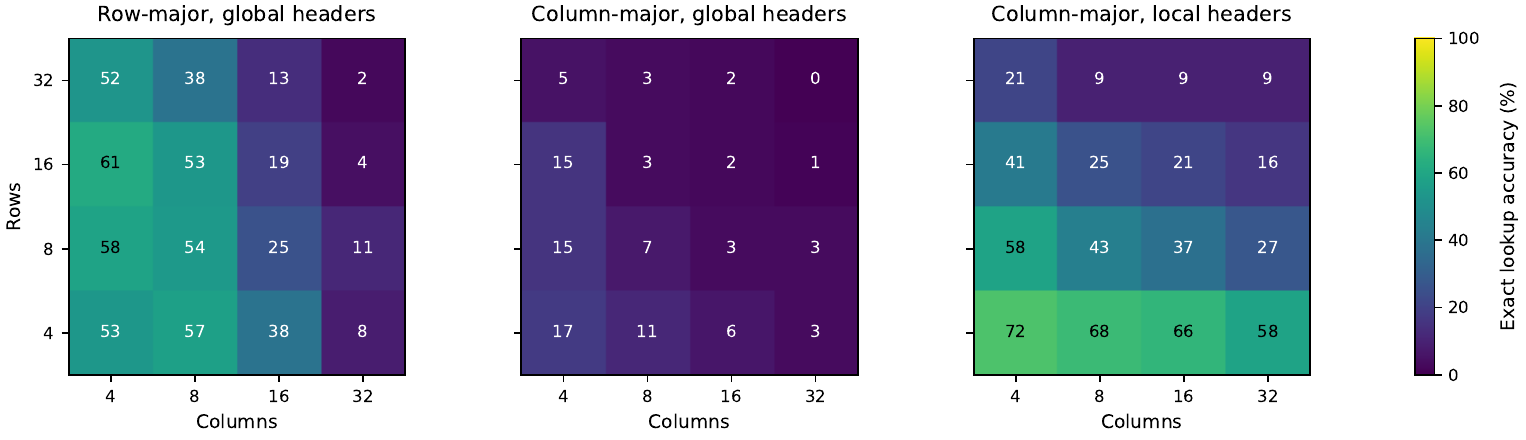}
\caption{Exact lookup accuracy in the independent-axis experiment (150 fresh
tables per cell). Transposing under global headers is uniformly poor. Attaching
each header to its column span rotates the vulnerable axis from columns to
rows.}
\label{fig:independent-grid}
\end{figure}

Figure~\ref{fig:independent-grid} reports all 16 shapes in the separately
generated grid. Pooled row-major/global-header, column-major/global-header, and
column-major/local-header accuracies are 34.0\%, 6.0\%, and 36.4\%,
respectively. The near-zero pooled difference between the two operational
formats ($+2.3$ points, CI $[-0.2,4.9]$) conceals the predicted crossover: at
$4\times32$, local column-major scores 58.0\% versus 8.0\% row-major, while at
$32\times4$, row-major scores 52.0\% versus 21.3\% local column-major.

\begin{table}[ht]
\centering
\small
\caption{Change in exact-lookup odds for each doubling of rows or columns in the independent-axis grid, with logistic-model 95\% intervals. Paired representation contrasts are reported separately.}
\label{tab:grid-slopes}
\begin{tabular}{lrr}
\toprule
Representation & Rows OR & Columns OR \\
\midrule
Row-major, global headers & 0.80 [0.74, 0.87] & 0.41 [0.37, 0.44] \\
Column-major, global headers & 0.65 [0.55, 0.77] & 0.48 [0.40, 0.58] \\
Column-major, local headers & 0.41 [0.37, 0.45] & 0.71 [0.65, 0.77] \\
\bottomrule
\end{tabular}
\end{table}

\FloatBarrier

An odds ratio below one means that doubling an axis reduces the odds of a
correct answer. Direct clustered interaction tests confirm that the operational
representations have different slopes for rows (interaction OR 0.51,
$p<10^{-26}$) and columns (OR 1.75, $p<10^{-18}$).

\section{Exact prompt templates}
Discovery row prompts declare a global column order and render one value list
per row. Global column prompts render one positional list per column; local
column prompts render \texttt{Column $H_j$: $v_{1j}$ | $\ldots$}. All end with
the exact question and answer-only instruction in
Figure~\ref{fig:prompt-example}.

Fragmentation prompts render a sequence of individually labeled
\texttt{GROUP=VALUE} records and ask: \texttt{Find the only value that occurs
in both group G1 and group G2.} Every condition repeats the same answer-only
instruction and 30-token response limit. Within a paired base instance, only
record order changes.

\section{Reproducibility details}
The constant-cell seed is 2026081801 and the independent-grid seed is
2026081502. The fragmentation dose response uses 20260825.
The Qwen Coder revision is
\texttt{488639f1ff808d1d3d0ba301aef8c11461451ec5}; Qwen 3B uses
\texttt{aa8e72537993ba99e69dfaafa59ed015b17504d1}; Phi-3.5 Mini uses
\texttt{2fe192450127e6a83f7441aef6e3ca586c338b77}. The Qwen checkpoints are
distributed under the Qwen Research License and Phi-3.5 Mini under the MIT
License; the pinned model cards are
\url{https://huggingface.co/Qwen/Qwen2.5-Coder-3B-Instruct},
\url{https://huggingface.co/Qwen/Qwen2.5-3B-Instruct}, and
\url{https://huggingface.co/microsoft/Phi-3.5-mini-instruct}.

All reported generation ran on RunPod RTX 4090 instances with 24\,GB memory.
The three crossover runs took 9, 12, and 17 minutes, and the independent grid
took 34 minutes. The ordering experiment ran on the same hardware; no training
was performed. A complete artifact with
generators, frozen manifests, exact prompts, raw responses, scoring outputs,
paired analyses, token summaries, plots, GPU metadata, and SHA-256 checksums
will be released after anonymous review.

\clearpage
\section*{NeurIPS Paper Checklist}

\begin{enumerate}

\item {\bf Claims}
    \item[] Question: Do the main claims made in the abstract and introduction accurately reflect the paper's contributions and scope?
    \item[] Answer: \answerYes{}
    \item[] Justification: The abstract and Sections 1--2 state both tasks, the tested model scope, paired interventions, principal effect sizes, and the deliberately behavioral interpretation.

\item {\bf Limitations}
    \item[] Question: Does the paper discuss the limitations of the work performed by the authors?
    \item[] Answer: \answerYes{}
    \item[] Justification: Section 4 limits the claims to synthetic inputs, two narrow diagnostic tasks, three-model crossover evidence, and a one-model non-tabular ordering experiment; it also states that the study identifies no mechanism and was not preregistered.

\item {\bf Theory assumptions and proofs}
    \item[] Question: For each theoretical result, does the paper provide the full set of assumptions and a complete (and correct) proof?
    \item[] Answer: \answerNA{}
    \item[] Justification: The paper reports behavioral experiments and statistical models, not theoretical results or proofs.

    \item {\bf Experimental result reproducibility}
    \item[] Question: Does the paper fully disclose all the information needed to reproduce the main experimental results of the paper to the extent that it affects the main claims and/or conclusions of the paper (regardless of whether the code and data are provided or not)?
    \item[] Answer: \answerYes{}
    \item[] Justification: Section 2 and Appendices A--E specify generation, scoring, paired statistics, prompt templates, seeds, checkpoint revisions, protocol deviations, and compute.

\item {\bf Open access to data and code}
    \item[] Question: Does the paper provide open access to the data and code, with sufficient instructions to faithfully reproduce the main experimental results, as described in supplemental material?
    \item[] Answer: \answerNo{}
    \item[] Justification: To preserve double-blind review for this single-PDF workshop submission, code and raw outputs are not linked in the review version; the complete artifact will be released after review.

\item {\bf Experimental setting/details}
    \item[] Question: Does the paper specify all the training and test details (e.g., data splits, hyperparameters, how they were chosen, type of optimizer) necessary to understand the results?
    \item[] Answer: \answerYes{}
    \item[] Justification: Sections 2.1--2.2 specify the sampled instances, conditions, prompt counts, models, decoding, and scoring; Appendices A and E provide output limits, pinned revisions, seeds, and hardware details. There is no training.

\item {\bf Experiment statistical significance}
    \item[] Question: Does the paper report error bars suitably and correctly defined or other appropriate information about the statistical significance of the experiments?
    \item[] Answer: \answerYes{}
    \item[] Justification: Sections 2.1--2.2 report Wilson 95\% intervals over independently sampled prompts and paired bootstrap intervals for representation contrasts. Appendix A specifies 10,000 paired resamples and exact McNemar tests; Appendix C reports the clustered interaction analyses used for the additional grid.

\item {\bf Experiments compute resources}
    \item[] Question: For each experiment, does the paper provide sufficient information on the computer resources (type of compute workers, memory, time of execution) needed to reproduce the experiments?
    \item[] Answer: \answerNo{}
    \item[] Justification: Appendix E reports the RunPod RTX 4090 type, 24\,GB memory, and durations for the three crossover runs and independent grid. The exact ordering-experiment duration and total development compute were not retained, so the disclosure is incomplete for this question's standard.
    
\item {\bf Code of ethics}
    \item[] Question: Does the research conducted in the paper conform, in every respect, with the NeurIPS Code of Ethics \url{https://neurips.cc/public/EthicsGuidelines}?
    \item[] Answer: \answerYes{}
    \item[] Justification: The study uses synthetic tables and synthetic labeled-record sequences, openly distributed model checkpoints under their stated terms, and no personal, private, or human-subject data.

\item {\bf Broader impacts}
    \item[] Question: Does the paper discuss both potential positive societal impacts and negative societal impacts of the work performed?
    \item[] Answer: \answerNA{}
    \item[] Justification: This is a controlled diagnostic of text serialization with no direct deployment; Section 4 explicitly recommends deterministic SQL or dataframe execution for exact production table retrieval.
    
\item {\bf Safeguards}
    \item[] Question: Does the paper describe safeguards that have been put in place for responsible release of data or models that have a high risk for misuse (e.g., pre-trained language models, image generators, or scraped datasets)?
    \item[] Answer: \answerNA{}
    \item[] Justification: The work releases neither a new model nor scraped or sensitive data, and the synthetic lookup artifacts do not create a material misuse risk.

\item {\bf Licenses for existing assets}
    \item[] Question: Are the creators or original owners of assets (e.g., code, data, models), used in the paper, properly credited and are the license and terms of use explicitly mentioned and properly respected?
    \item[] Answer: \answerYes{}
    \item[] Justification: Section 2 cites the model reports, and Appendix E gives pinned revisions, model-card URLs, and the Qwen Research and MIT licenses.

\item {\bf New assets}
    \item[] Question: Are new assets introduced in the paper well documented and is the documentation provided alongside the assets?
    \item[] Answer: \answerNA{}
    \item[] Justification: No new dataset or model is released with the anonymous submission; the paper describes a synthetic experimental artifact that will be released after review.

\item {\bf Crowdsourcing and research with human subjects}
    \item[] Question: For crowdsourcing experiments and research with human subjects, does the paper include the full text of instructions given to participants and screenshots, if applicable, as well as details about compensation (if any)? 
    \item[] Answer: \answerNA{}
    \item[] Justification: The research contains no crowdsourcing or human-subject participation.

\item {\bf Institutional review board (IRB) approvals or equivalent for research with human subjects}
    \item[] Question: Does the paper describe potential risks incurred by study participants, whether such risks were disclosed to the subjects, and whether Institutional Review Board (IRB) approvals (or an equivalent approval/review based on the requirements of your country or institution) were obtained?
    \item[] Answer: \answerNA{}
    \item[] Justification: The research contains no human subjects and therefore requires no IRB or equivalent review.

\item {\bf Declaration of LLM usage}
    \item[] Question: Does the paper describe the usage of LLMs if it is an important, original, or non-standard component of the core methods in this research? Note that if the LLM is used only for writing, editing, or formatting purposes and does \emph{not} impact the core methodology, scientific rigor, or originality of the research, declaration is not required.
    \item[] Answer: \answerYes{}
    \item[] Justification: Language-model inference is the object of study; Section 2 identifies the evaluated checkpoints and describes prompting, decoding, and scoring, with full details in Appendices A and E.

\end{enumerate}

\end{document}